# Mathematical Modeling Of Four Finger Robotic Grippers

Sajjad Hussain[1], M. Suhaib[1]

[1]Department of Mechanical Engineering, Jamia Millia Islamia, New Delhi, 110025 India

**Abstract**

Robotic grippers are the end effector in the robot system of handling any task which used for performing various operations for the purpose of industrial application and hazardous tasks.In this paper, we developed the mathematical model for multi fingers robotics grippers. we are concerned with Jamia'shand which is developed in Robotics Lab, Mechanical Engineering Deptt, Faculty of Engg & Technolgy, Jamia Millia Islamia, India. This is a tendon-driven gripper each finger having three DOF having a total of 11 DOF. The term tendon is widely used to imply belts, cables, or similar types of applications. It is made up of three fingers and a thumb. Every finger and thumb has one degree of freedom. The power transmission mechanism is a rope and pulley system. Both hands have similar structures. Aluminum from the 5083 families was used to make this product. The gripping force can be adjusted we have done the kinematics, force, and dynamic analysis by developing a Mathematical model for the four-finger robotics grippers and their thumb. we focused it control motions in X and Y Displacements with the angular positions movements and we make the force analysis of the four fingers and thumb calculate the maximum weight, force, and torque required to move it with mass. Draw the force -displacements graph which shows the linear behavior up to 250 N and shows nonlinear behavior beyond this. and required $D_{min}$ of wire is 0.86 mm for grasping the maximum 1 kg load also developed the dynamic model (using energy )approach lagrangian method to find it torque required to move the fingers.

1. Introduction

An industrial robot is a reprogrammable multipurpose manipulator that can move material, parts, tools, and specialized devices using varied programmed motions to execute a wide range of activities. Robotics Grippers is one of the Important Parts Of Robot Handling Of Any Object or workpiece for performing various Engineering .tasks Or For Industrial & Hazardous work, we use various types of Grippers and Their actuation mechanism. so we are concerned about Multi Fingers Robotics Grippers Which Consist of Three Similar Finger. There are three fingers the first finger is known as proximal, the second is middle, and the third is distal phalanges are respectively, and the one is the thumb. And this three-finger and one thumb hand are developed in the Robotics Lab, Mechanical Engineering Deptt, Faculty of Engg& Technolgy, Jamia Millia Islamia. This is a tendon-driven gripper each finger having three DOF having a total of 11 DOF.[1]. The term tendon is widely used to imply belts, cables, or similar types of applications. Specifically, this work is concerned with the systematic approaches for the synthesis of tendon routing in a robotic manipulator from the kinematic &dyamic point of view, rather than from the geometric or material point of view. Experience with tendons has pointed out that tendons are commonly used to transmit power between shafts where center distance is to great for gearing or similar power-transmission devices.

## 2. Literature Review

Ion Simionescu and Ion Ion[2]. focused on the kinematics part of the grippers and its location of moving because the Grippers are the final link in the kinematic chain in industrial robot joint systems, allowing for easier engagement with various task its surrounding . there are multipurpose grippers with enormous holding capacities can be utilized for a variety of workpiece shapes, they must be customized in many circumstances. Tatiana Victorovna Zudilova et al.[3]. the researchers demonstrated the mathematical modeling of the four fingers manipulator and want to transfer its analogy to the four fingers robotics grippers that end effector for the robot system that supports a robot mathematical modeling design. Robot manipulators are utilized in a variety of fields and realms of human activity, including assembling, machining, and heat treatment. Kinga Stasik[4] . presents the dynamics of a pinch-type gripper for textiles objects a gripper in which the closing and opening motion is controlled by a screw. A set of equations defining the gripper's behavior has been numerically solved, preceded by a differential equation of the driving torque of the gripper's driving motor. utilizing the Runge-Kutta method suited for the best-manipulated method. The mathematical model allows gripper parameters to be selected during the design stage. The gripper's force on a cloth is determined by a user-accepted attribute.Ehtesham Nazma and Suhaib Mohd [5] demonstrated that in the robotics grippers It is frequently required to employ a transmission system that allows the actuators to be positioned away from the site of application to reduce the inertia of a manipulator. E. Nehaa et al [6] explore the Computations of grasp analysis for the four-finger robotics grippers. The tendon wires and spring are used in a four-finger robotic hand. Any multi-finger robotic hand should be capable of stable grasping and delicate manipulation of a variety of objects. Various shape like cylinder,sphere performance of the four-fingered robotic evalatuted on the sim mechanics MATLAB .

Megan Grimm et al.[7] present a novel 3 link robotic manipulator which is used for gymnastics purposes develop the dynamic mathematical corresponding to each link of the manipulator and draw its trajectory.Rituparna Datta et al [8] present the study about a nonlinear, multimodal, and multiobjective optimization problem that was conceived The prior work, on the other hand, had dealt with As a black box, consider the actuator. The system model has been altered by the addition of an actuarial component. Each of the nondominated solutions yield force voltage relationship, which aids the user in determining the voltage to apply based on the application.

## 3. CAD Model Of Multi-Finger Robotics Grippers

The proposed robot hand's solid model has four fingers, three of which are in series and one of which is in opposition to the center of the three series. There are three of them. Each finger has phalanges links. The joint in the palm link is fixed. and think about being the foundation. One is present in both the median and distal connections. Rotational joints with a degree of freedom (DOF). The material chosen is aluminum because of its lightweight and good strength. Figure 1 shows the CAD model of the gripper.

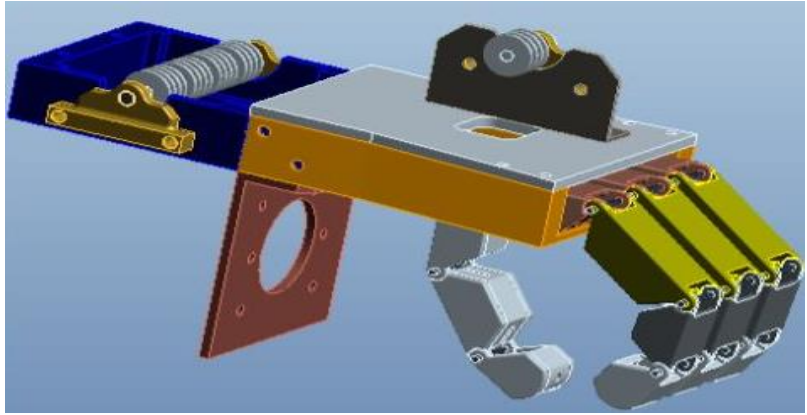

**Figure 1:** CAD model of four fingers Jamia Hand [1]

### 4. Kinematics Modeling Of The Four-Finger Gripper

Kinematics Modeling generally focused on the gripper displacement with respect to its angular positions .and draw its possible reach-out path during operating. L1 is the length of the link that is known as proximal phalanges, link 2 is intermediate phalanges, and link 3 is distal phalanges. Figure 2 indicate the kinematic model of the gripper.

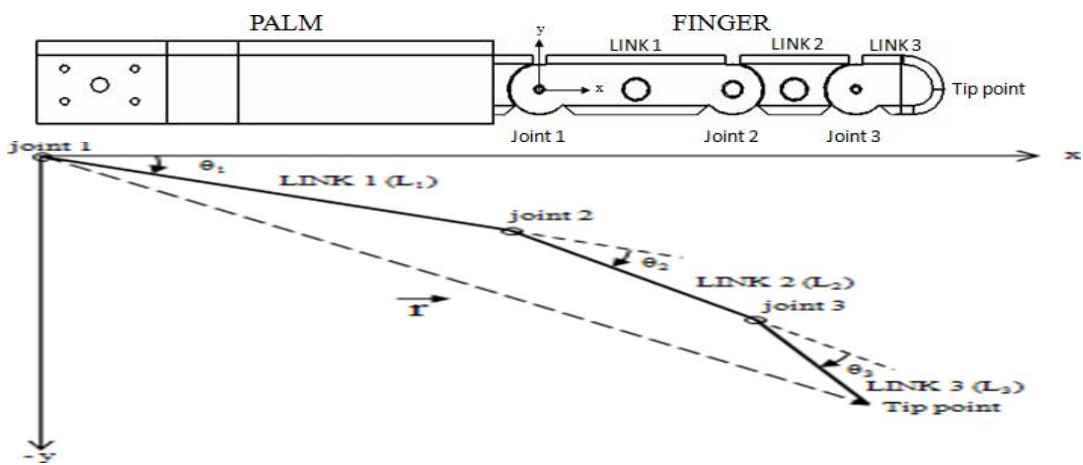

**Figure 2:** Gripper finger and its kinematics model [9]

The movement of the three finger can be seen from their kinematics modeling ,in this model we have observed the rotation of the each finger link joint at each point is about z axis and other two x and y axix are locked and the rotation angle of first to third link are always equal. So we can write the position vector of combining each finger movement about z axix rotation For finger link $L_1$ Movement in X axis $L_1 \cos \theta_1$, for link 2, $L_2 \cos (\theta_1 + \theta_2)$ and for the link three $L_3 \cos (\theta_1 + \theta_2 + \theta_3)$ and similarly in Y axis for link L1 is $L_1 \sin \theta_1$ for link $L_2$ is $L_2 \sin (\theta_1 + \theta_2)$ and for link three is $L_3 \sin (\theta_1 + \theta_2 + \theta_3)$, for the final movement in X and Y directions of the link putting the value of three different link length $L_1$= 30 mm  length $L_2$= 15 mm length $L_3$= 10 mm respectively .and assume all three angular movement $\theta1 = \theta2 = \theta3 = \theta$  are equal. Figure 3 indiacte the X-Y displacement with angular rotaion of the gripper.

$$X = 30 \cos \theta + 15 \cos 2\theta + 10 \cos 3\theta \qquad (1)$$

$$Y = 30 \sin \theta + 15 \sin 2\theta + 10 \sin 3\theta \qquad (2)$$

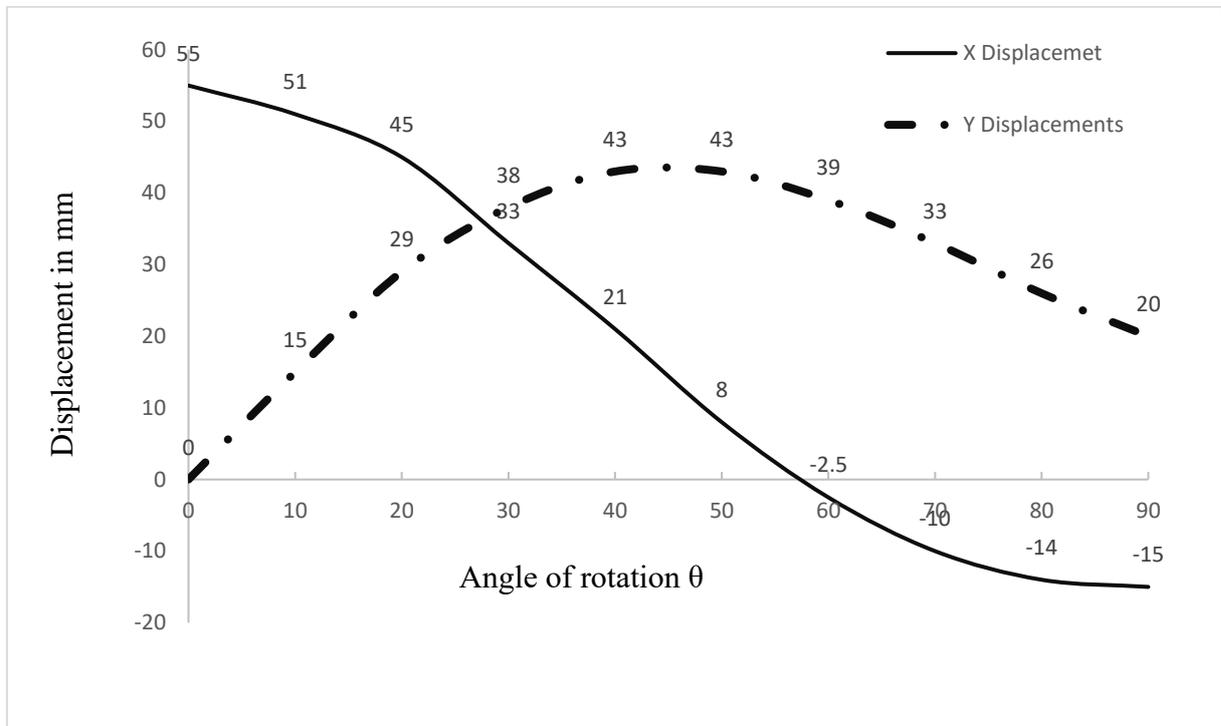

**Figure 3:** X-Y displacements with angular rotation

### 5. Force Analysis On The Fingers

Because the other two fingers have an identical design, only one finger is subjected to force analysis. The first step in performing a force analysis is to construct a schematic model of the finger.

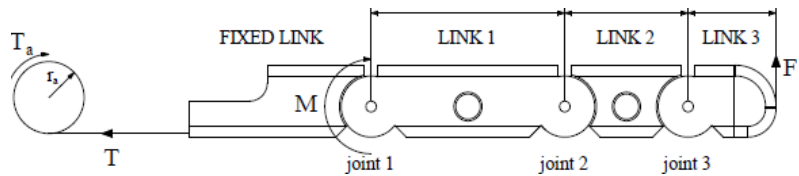

**Figure 4:** Finger representation

$L_1$ is the length of the link that is known as proximal phalange link2 is intermediate phalange and link 3 is distal phalanges moment at joint

$M_1 = F * (L_1 + L_2 + L_3)$      $M_1 = 55F$ Moment at first three joint
$M_2 = F * (L_2 + L_3)$      $M_2 = 25F$ Moment at first two joints from the end of the link
$M_3 = F * (L_3)$      $M_3 = 10F$ Moment at first joint at the end link
$M_1$ = Maximum moment and $M = T * r_p$ Where T is the maximum tension force generated in the wire of the rope and radius of the pulley is denoted by $r_P$ at joint Takes $r_{p=}$ 3 mm 55F=T*3 T=18.33F Now Torque required to rotate    $T_{orque} = T * r_a$, radius of the actuator ($r_a$ =5mm ) $T_{orque} = 18.33F * 5 * 10^{-3}$, $T_{orque} = 0.09165F$ [N-m], Max Gripping Force =6 N
Maximum carrying capacity weight, $m = F * \mu / g$   m= mass   u = coefficient of friction
$\mu = 0.5$   g=9.81m/s²   m= 6*0.5/9.81 = 0.305 kg

so similar to three fingers we can lift m =3*0.305=0.915 kg

Stress Calculation $\sigma = \frac{T}{A}$    $A=\pi d2/4$    $\sigma_{all}=190$ Mpa

$D_{min}= \sqrt{\frac{4T}{\pi \sigma_a}}$ =0.866 mm

During the gripping of the of maximum load of the object we have observed the maximum tension and observed the linear behavior up to 200 N beyond this showing the non linear behavior, in the force displacement curve. The Figure 5 force and displacement graph.

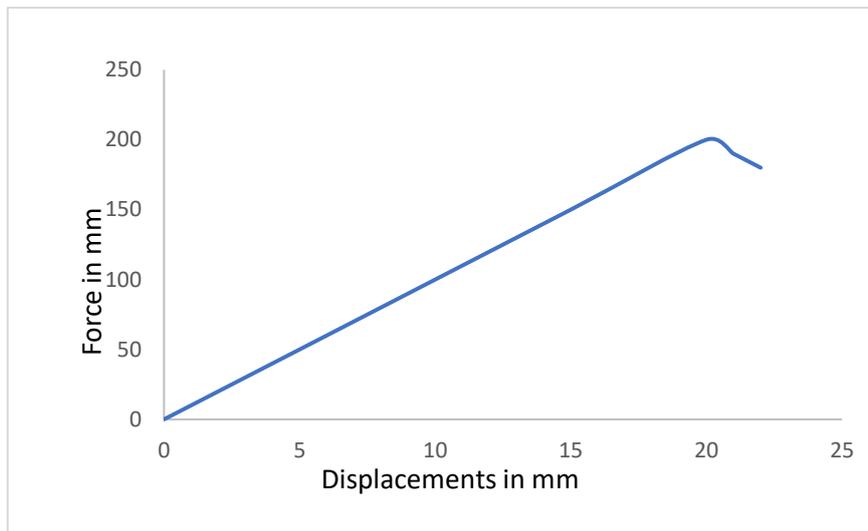

**Figure 5:** force-displacements grap

## 6. Force Anslysis of Thumb

The Moment Equations at Joint:

$M_{T1} = F_T * (L_1 + L_2)$      $M_{T1} = 25F_T$   Moment at each joint from the end
$M_{T2} = F_T * (L_2)$             $M_{T2} = 10F_T$   Moment at end joint link2
Where $L_1 = 15$ mm , $L_2 = 10$ mm

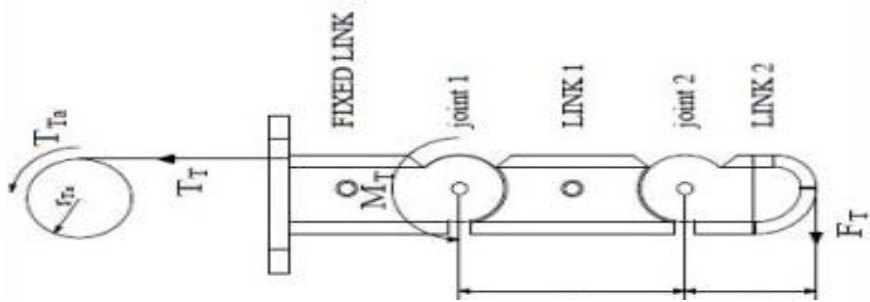

**Figure 6:** Thumb representation

$M_1$ = Maximum moment and $M = T * r_p$ Where T is the maximum tension force generated in the wire of the rope and radius of the pulley is denoted by $r_P$ at joint Takes
Takes $r_p=$ mm , 25F=T* , T=8.33F
Torque required for rotation,   $T_{orque} = T*r_a$ , where radius of the actuator ($r_a$) =5mm
$T_{orque} = 8.33F*5*10^{-3}$, $T_{orque} = 0.04165F$ [N-m] · Max Gripping Force =10 N , $T_{orque} = 0.4166$ [N-m]

## 7. Dynamic Modelling Using Lagrangian Method

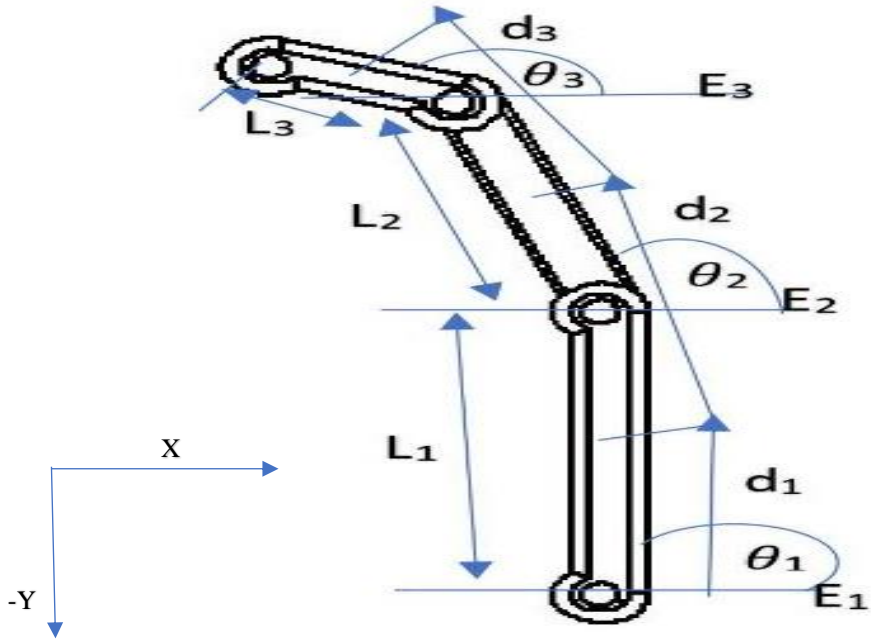

**Figure 7:** Different joints in the finger link

Lagrangian method is the Energy- based approach in which we have write the total potential energy of each fingers and total kinetic energy of each fingers using different fingers terminology Each of the $E_1$, $E_2$ and $E_3$ the joint is revolute [10] . Each of the phalanges makes $\theta_1$ , $\theta_2$, and $\theta_3$ absolute angles with the x-axis, correspondingly. Let The lengths of the phalanges $L_1, L_2$ are $L_3$ , and p3. Let $d_1$ be an additional variable. $d_2$ , $d_3$ are the distances between the phalanges' centers of mass. $E_1$, $E_2$ and $E_3$ , as depicted in the diagram, come from their respective joints. figure. Let $I_1$, $I_2$ and $I_3$ are the movement of inertia of each finger of the link along the - y-axis Finally, make $m_1, m_2$ and $m_3$ the masses. The proximal, middle and distal phalanges are respectively. Figure 7 shows the three finger joint.

The kinetic energy of the first fingers link (proximal), $K_1$

$$K_1 = \tfrac{1}{2} I_1 \dot{\theta}_1^{\,2} + \tfrac{1}{2} m_1 d_1^2 \dot{\theta}_1^2 \qquad (3)$$

The kinetic energy of the second link (middle phalange) , $K_2$

$$K_2 = \tfrac{1}{2} I_2 \dot{\theta}_2^2 + \tfrac{1}{2} m_2 \{L_1^2 \dot{\theta}_1^2 + d_2 \dot{\theta}_2^2 + 2 L_1 d_2 \dot{\theta}_2 \dot{\theta}_1 \cos(\theta_1 - \theta_2)\} \qquad (4)$$

The kinetic energy of the third link ( Distal phalange) ,$K_3$

$$K_3 = \tfrac{1}{2} I_3 \dot{\theta}_3^2 + \tfrac{1}{2} m_3 \{L_1^2 \dot{\theta}_1^2 + L_2 \dot{\theta}_2^2 + d_3^2 + \ 2[L_1 L_2 \dot{\theta}_1 \dot{\theta}_2 \cos(\theta_1 - \theta_2) + L_2 d_2 \dot{\theta}_2 \dot{\theta}_3 \cos(\theta_2 - \theta_3)] + L_1 d_3 \dot{\theta}_1 \dot{\theta}_3 \cos(\theta_1 - \theta_3) \} \qquad (5)$$

Hence, adding all three kinetic energy ie total kinetic energy of the three link $K = K_1 + K_2 + K_3$
And total potential energy of three link

$$P = m_1 g d_1 \sin \theta_1 + m_2 g(L_1 \sin \theta_1 + d_2 \sin \theta_2) \\ + m_2 g[L_1 \sin \theta_1 + L_2 \sin \theta_2\, d_3 \sin \theta_3 \ ] \qquad (6)$$

Applying the lagrangian concept ie the difference of the total kinetic energy of the link to the to total potential energy of the link

$L(\theta_1, \theta_2, \theta_3, \dot{\theta}_1, \dot{\theta}_2, \dot{\theta}_3) = K(\theta_1, \theta_2, \theta_3, \dot{\theta}_1, \dot{\theta}_2, \dot{\theta}_3) - P(\theta_1, \theta_2, \theta_3)$

Let the different torque are $T_1$, $T_2$, and $T_3$ at each joint $E_1$, $E_2$ and $E_3$ be respectively. Using three Euler lagrangian differential equations for finding the torque at each three joint

$$\frac{d}{dt}\left(\frac{\partial L}{\partial \dot{\theta}_1}\right) - \frac{\partial L}{\partial \theta_1} = T_1 \tag{7}$$

$$\frac{d}{dt}\left(\frac{\partial L}{\partial \dot{\theta}_2}\right) - \frac{\partial L}{\partial \theta_2} = T_2 \tag{8}$$

$$\frac{d}{dt}\left(\frac{\partial L}{\partial \dot{\theta}_3}\right) - \frac{\partial L}{\partial \theta_3} = T_3 \tag{9}$$

Using the above equations to finding the torque at each joint

$$T_1 = [I_1 + m_1 d_1^2 + m_2 L_1^2]\ddot{\theta}_1 + [L_1 d_2 + L_L L_2]\cos(\theta_1 - \theta_2)\ddot{\theta}_2 + L_1 d_3 \cos(\theta_1 - \theta_3)\ddot{\theta}_3 - [L_2 d_2 + L_1 L_2]\sin(\theta_1 - \theta_2)(\dot{\theta}_1 - \dot{\theta}_2)\dot{\theta}_2 - [a\ d_1 + m_2 L_1 + m_3 L_3]\ g\cos\theta_1 \tag{10}$$

$$T_2 = [I_2 + m_2 d_2^2 + m_3 L_2^2]\ddot{\theta}_2 + [m_2 d_2 + m_3 L_2] L_1 \cos(\theta_1 - \theta_2)\ddot{\theta}_1 + m_3 L_2 d_2 \cos(\theta_2 - \theta_3)\ddot{\theta}_3\ [m_2 d_2 + m_3 L_2] L_1 \sin(\theta_1 - \theta_2)(\dot{\theta}_1 - \dot{\theta}_2)\dot{\theta}_1 - m_3 L_2 d_3 \sin(\theta_2 - \theta_3)(\dot{\theta}_2 - \dot{\theta}_3)\dot{\theta}_3 + [m_2 d_2 + m_3 L_2] L_1 \sin(\theta_1 - \theta_2)\dot{\theta}_1 \dot{\theta}_2 + L_2 d_3 \sin(\theta_2 - \theta_3)\dot{\theta}_2 \dot{\theta}_3 + [m_2 d_2 + m_3 L_2] g\cos\theta_2 \tag{11}$$

$$T_3 = [I_3 + m_3 d_3^2]\ddot{\theta}_3 + [m_3 L_2 d_3 \cos(\theta_2 - \theta_3)]\ddot{\theta}_2 + m_3 L_1 d_3 \cos(\theta_1 - \theta_3)]\ddot{\theta}_1 - [m_3 L_2 d_3 \sin(\theta_2 - \theta_3)(\dot{\theta}_2 - \dot{\theta}_3)\dot{\theta}_2 + m_3 L_1 d_3 \sin(\theta_1 - \theta_3)\dot{\theta}_1 \dot{\theta}_3 + L_2 d_3 \sin(\theta_2 - \theta_3)\dot{\theta}_2 \dot{\theta}_3 + + m_3 d_3\ g\cos\theta_3 \tag{12}$$

8. **Conclusions**

Robotics grippers is End-effectors which is important in robotic applications since they are the portions of the robot that come into direct touch with objects. End-effectors such as grippers are utilised to manipulate work components. Grippers come in a variety of shapes and sizes,and its actuation mechanism choosing the proper one for the job is vital. We are concern with to choose the varstile grippers which is used for the multi purpose of garsping the object . Special purpose grippers, are more flexible and stable than. Multi fingers grippers have greater dexterity than others grippers, and their carrying capacity is wide range . Both specific task grippers and multi-purpose grippers are investigated in this thesis. developed. As its duty is defined, a special task gripper is built with four fingers. Both a special task gripper and a multi-purpose gripper are created in this thesis. Because the mission of the special task gripper is established at the start of the design process, it is designed with two fingers. As a result, needs are easily established, and design is simplified. The procedure progressed quite quickly. The multipurpose gripper, on the other hand, had no predetermined task. As a result, past instances and the human hand are employed. we developed the mathematical model for the MFRG grippers using the lagrangian concept. the force controlling is calculated and developed the kinematics mathematical and shown the X Y displacements with angular movements of the positions. Dynamic modeling suggested the approximate value of the torque which depend on the various parameter of the four fingers robotics grippers its mass center distance, a moment of inertia and its angular movement.